\documentclass[10pt,twocolumn,letterpaper]{article}

\usepackage{cvpr}
\usepackage{times}
\usepackage{epsfig}
\usepackage{graphicx}
\usepackage{amsmath}
\usepackage{amssymb}
\usepackage{algorithm} 
\usepackage{algorithmic}
\usepackage{helvet}
\usepackage[english]{babel}
\usepackage{courier}
\usepackage{bm}
\usepackage[marginal]{footmisc}

\usepackage{indentfirst}
\usepackage{multirow}
\usepackage{blindtext}
\usepackage{color}
\usepackage{booktabs}
\usepackage{setspace}

\usepackage{balance}
\usepackage{booktabs}
\usepackage{setspace}

\usepackage[pagebackref=true,breaklinks=true,letterpaper=true,colorlinks,bookmarks=false]{hyperref}
\makeatletter

\cvprfinalcopy 

\begin{document}

\title{FastReID: A Pytorch Toolbox for General Instance Re-identification} 
\author{Lingxiao He$^{* \dagger}$, Xingyu Liao$^{*}$, Wu Liu$^{\dagger}$, Xinchen Liu, Peng Cheng and Tao Mei \\
JD AI Research\\
{\tt\small \{helingxiao3, liaoxingyu5, liuwu1, liuxinchen1, chengpeng8, tmei\}@jd.com}
}

\maketitle

\begin{abstract}

$\footnote{*{Authors contributed equally}; $\dagger${Corresponding author}\\
\textbf{Acknowledgements} Thanks to Kecheng Zheng (\emph{zkcys001@mail.ustc.edu. cn}),
Jinkai Zheng (\emph{zhengjinkai3@hdu.edu.cn}) and Boqiang Xu (\emph{boqiang.xu@cripac.ia.ac.cn}), partial works were done when they interned at JD AI Research.}$
General Instance Re-identification is a very important task in the computer vision, which can be widely used in many practical applications, such as person/vehicle re-identification, face recognition, wildlife protection, commodity tracing, and snapshop, etc.. To meet the increasing application demand for general instance re-identification, we present FastReID as a widely used software system in JD AI Research. In FastReID, highly modular and extensible design makes it easy for the researcher to achieve new research ideas. Friendly manageable system configuration and engineering deployment functions allow practitioners to quickly deploy models into productions. We have implemented some state-of-the-art projects, including person re-id, partial re-id, cross-domain re-id and vehicle re-id, and plan to release these pre-trained models on multiple benchmark datasets. FastReID is by far the most general and high-performance toolbox that supports single and multiple GPU servers, you can reproduce our project results very easily and are very welcome to use it, the code and models are available at https: \url{https://github.com/JDAI-CV/fast-reid}.

\end{abstract}
\section{Introduction}
General instance re-identification (re-id), as an instance-centric AI technique, aiming at finding a certain person/vehicle/face/object of interest in a large amount of videos. It facilitates various applications that require painful and boring video watching, including searching for video shots related to an actor of interest from TV series, a lost child in a shopping mall from camera videos, a suspect vehicle from a city surveillance system. Moreover, the General instance re-identification technique is also used for snapshop in e-commerce platforms, commodity tracing in merchandise security and wildlife protection. Many researchers realize a task based on open source code, less extensible and reusable modification make it difficult to reproduce the results. Besides, there often exists a gap between academic research and practical applications, which makes it difficult for academic research techniques to be quickly transferred to productions.

\begin{figure*}[t]
    \centering
       \vspace{0em}
    \includegraphics[width=16cm]{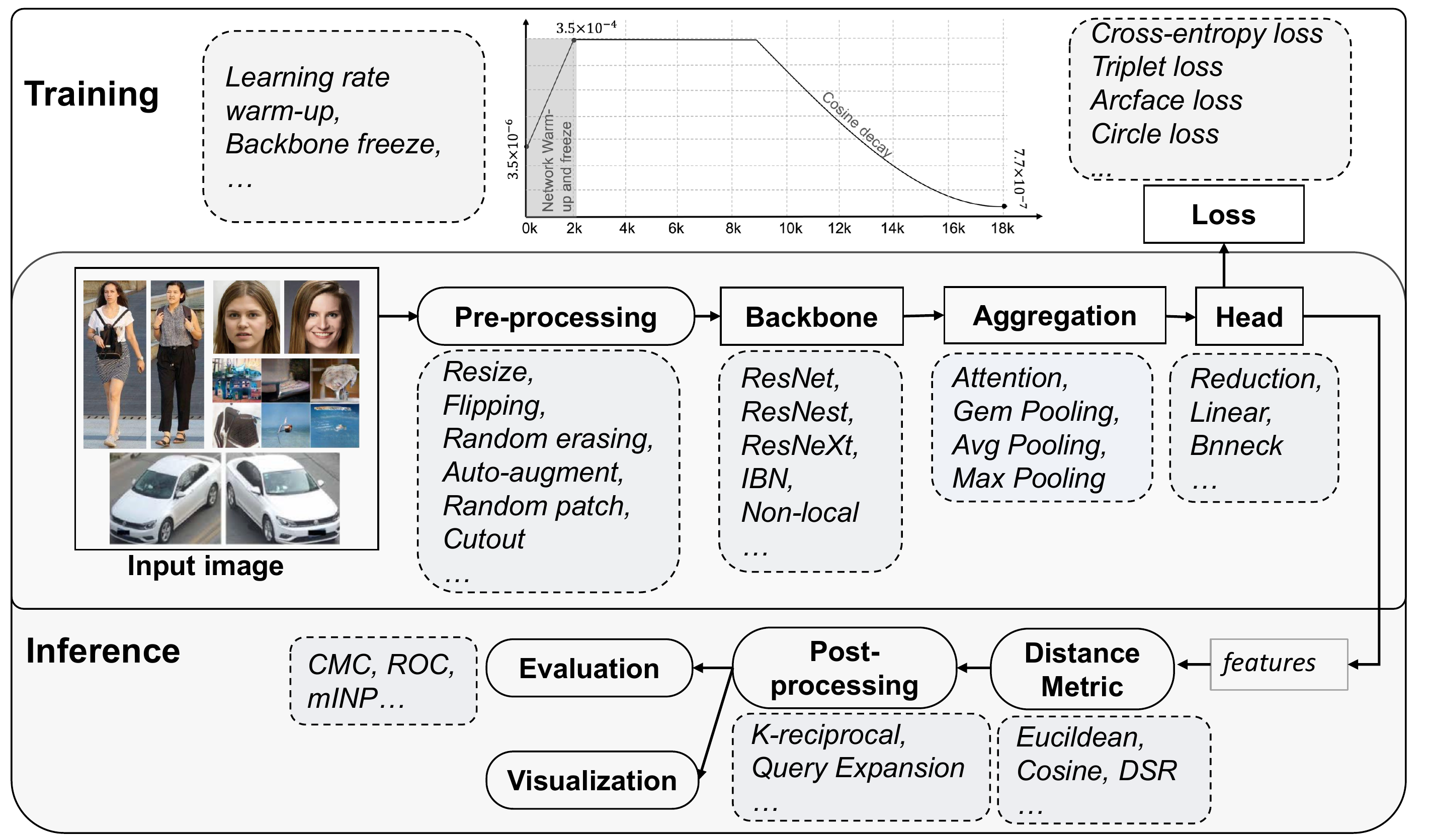}
     \caption{The Pipeline of FastReID library.}
    \label{fig1}
    \vspace{-1em}
\end{figure*}

To accelerate progress in the community of general instance re-identification including researchers and practitioners in academia and industry, we now release a unified instance re-id library named FastReID. We have introduced a stronger modular, extensible design that allows researchers and practitioners easily to plug their oven designed module without repeatedly rewriting codebase, into a re-id system for further rapidly moving research ideas into production models. Manageable system configuration makes it more flexible and extensible, which is easily extended to a range of tasks, such as general image retrieve and face recognition, etc. Based on FastReID, we provide many state-of-the-art pre-trained models on multiple tasks about person re-id, cross-domain person re-id, partial person re-id and vehicle re-id, and in the future we will release face recognition and object retrieval models. Besides, we hope that the library can provide a fair comparison between different approaches. 

Recently, FastReID has become one of the widely used open-source library in JD AI Research. We will continually refine it and add new features to it. We warmly welcome individuals, labs to use our open-source library and look forward to cooperating with you to jointly accelerate AI Research and achieve technological breakthroughs.

\section{Highlight of FastReID}
FastReID provides a complete toolkit for training, evaluation, finetuning and model deployment. Besides, FastReID provides strong baselines that are capable of achieving state-of-the-art performance on multiple tasks.

\noindent\textbf{Modular and extensible design.} In FastReID, we introduce a modular design that allows users to plug custom-designed modules into almost any part of the re-identification system. Therefore, many new researchers and practitioners can quickly implement their ideas without re-writing hundreds of thousands of lines of code.

\noindent\textbf{Manageable system configuration.} FastReID implemented in PyTorch is able to provide fast training on multi-GPU servers. Model definitions, training and testing are written as YAML files. FastReID supports many optional components, such as backbone, head aggregation layer and loss function, and training strategy.

\noindent\textbf{Richer evaluation system.} At present, many researchers only provide a single CMC evaluation index. To meet the requirement of model deployment in practical scenarios, FastReID provides more abundant evaluation indexes, e.g., ROC and mINP, which can better reflect the performance of models.

\noindent\textbf{Engineering deployment.} Too deep model is hard to deploy in edge computing hardware and AI chips due to time-consuming inference and unrealizable layers. FastReID implements the knowledge distillation module to obtain a more precise and efficient lightweight model. Also, FastReID provides a conversion tool, e.g., PyTorch$\to$Caffe and PyTorch$\to$TensorRT to achieve fast model deployment.

\noindent\textbf{State-of-the-art pre-trained models.} FastReID provides state-of-the-art inference models including person re-id, partial re-id, cross-domain re-id and vehicle re-id. We plan to release these pre-trained models. FastReID is very easy to extend to general object retrieval and face recognition. We hope that a common software advanced new ideas to applications.

\section{Architecture of FastReID}

In this section, we elaborate on the pipeline of FastReID as shown in Fig.~\ref{fig1}. The whole pipeline consists of four modules: image pre-processing, backbone, aggregation and head, we will introduce them in detail one by one.

\begin{figure}[t]
    \centering
       \vspace{0em}
    \includegraphics[width=7cm]{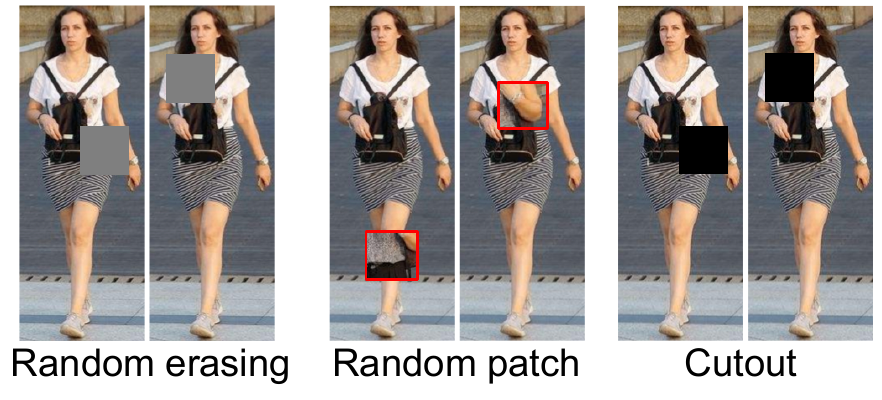}
     \caption{Image pre-processing.}
    \label{fig2}
    \vspace{-1em}
\end{figure}
\subsection{Image Pre-processing}
The collected images are of different sizes, we first resize the images to fixed-size images. And images can be packaged into batches and then input into the network. To obtain a more robust model, \textbf{flipping} as a data augmentation method by mirroring the source images to make data more diverse.  \textbf{Random erasing}, \textbf{Random patch} \cite{Zhou_2019_ICCV} and \noindent\textbf{Cutout} \cite{devries2017improved} are also augmentation methods that randomly selects a rectangle region in an image and erases its pixels with random values, another image patch and zero values, making the model effectively reduce the risk of over-fitting and robust to occlusion. \noindent\textbf{Auto-augment} is based on automl technique to achieve effective data augmentation for improving the robustness of feature representation. It uses an auto search algorithm to find the fusion policy about multiple image processing functions such as translation, rotation and shearing.

\subsection{Backbone}
Backbone is the network that infers an image to feature maps, such as a ResNet without the last average pooling layer. FastReID achieves three different backbones including ResNet \cite{he2016deep}, ResNeXt \cite{xie2017aggregated} and ResNeSt \cite{zhang2020resnest}. We also add attention-like non-local \cite{wang2018non} module and instance batch normalization (IBN) \cite{pan2018two} module into backbones to learn more robust feature. 

\subsection{Aggregation}
The aggregation layer aims to aggregate feature maps generated by the backbone into a global feature. We will introduce four aggregation methods: max pooling, average pooling, GeM pooling and attention pooling. The pooling layer takes $\mathbf{X}\in \mathbb{R}^{W\times H\times C}$ as input and produces a vector $\mathbf{f}\in \mathbb{R}^{1\times1\times C}$ as an output of the pooling process, where $W, H, C$ respectively represent the width, the height and the channel of the feature maps. The global vector $\mathbf{f} = [f_1,...,f_c,...,f_C]$ in the case of the max pooling, average pooling, GeM pooling and attention pooling of are respectively given by 
\begin{equation}
 \text{Max Pooling}: f_c=\max_{x\in \mathbf{X}_c}x
\label{eq1}
\end{equation}

\begin{equation}
 \text{Avg Pooling}: f_c= \frac{1}{|\mathbf{X}_c|}\sum_{x\in \mathbf{X}_c}x
\end{equation}

\begin{equation}
 \text{Gem Pooling}: f_c= (\frac{1}{|\mathbf{X}_c|}\sum_{x\in \mathbf{X}_c} x^{\alpha})^{\frac{1}{\alpha}}
\end{equation}

\begin{equation}
\text{Attention Pooling}: f_c= \frac{1}{|\mathbf{X}_c*\mathbf{W}_c|}\sum_{x\in \mathbf{X}_c, w\in \mathbf{W}_c} w*x
\end{equation}
where $\alpha$ is control coefficient and $\mathbf{W}_c$ are the softmax attention weights.

\subsection{Head}
\begin{figure}[t]
    \centering
       \vspace{0em}
    \includegraphics[width=8cm]{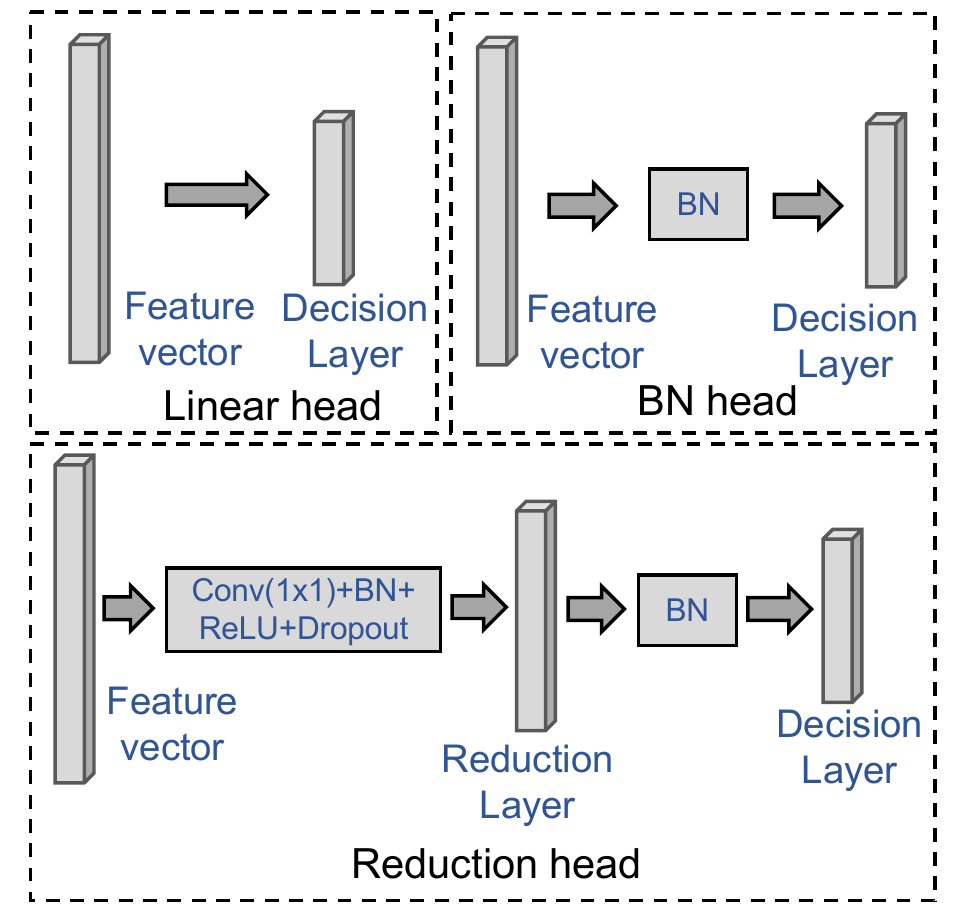}
     \caption{Different heads that implemented in FastReID}
    \label{fig4}
\end{figure}

Head is the part of addressing the global vector generated by aggregation module, including batch normalization (BN) head, Linear head and Reduction head. Three types of the head are shown in Fig.~\ref{fig4}, the linear head only contains a decision layer, the BN head contains a bn layer and a decision layer and the reduction head contains conv+bn+relu+dropout operation, a reduction layer and a decision layer. 

\noindent\textbf{Batch Normalization} \cite{ioffe2015batch} is used to solve internal covariate shift because it is very difficult to train models with saturating non-linearities. Given a batch of feature vector $\mathbf{f} \in \mathbb{R}^{m \times C}$ ($m$ is the sample number in a batch), then the bn feature vector $\mathbf{f}_{bn} \in \mathbb{R}^{m \times C}$ can be computed as

\begin{equation}
\left\{\begin{aligned}    
\mu & = \frac{1}{m}\sum_{i=1}^{m} \mathbf{f}_i, \\ 
\sigma^{2} & = \frac{1}{m}\sum_{i=1}^{m} (\mathbf{f}_i-\mu)^2, \\
\mathbf{f}_{bn} & = \gamma\cdot\frac{\mathbf{f}-\mu}{\sqrt{\sigma^{2}+\epsilon}}+\beta
 \end{aligned}\right.
\end{equation}
where $\gamma$ and $\beta$ are trainable scale and shift parameters, and $\epsilon$ is a constant added to the mini-batch variance for numerical stability.

\noindent\textbf{Reduction layer} is aiming to make the high-dimensional feature become the low-dimensional feature, i.e., 2048-dim$\to$512-dim.

\noindent\textbf{Decision layer} outputs the probability of different categories to distinguish different categories for the following model training.
\section{Training}
\subsection{Loss Function}
Four different loss functions are implemented in FastReID. 

\noindent\textbf{Cross-entropy loss} is usually used for one-of-many classification, which can be defined as
\begin{equation}
 \mathcal{L}_{ce} = \sum_{i=1}^{C}y_i\log \hat{y_i} + (1-y_i)\log(1-\hat{y_i}),
\end{equation}
where $\hat{y_i}=\frac{e^{\mathbf{W}_i^{T}\mathbf{f}}}{\sum_{i=1}^{C}e^{\mathbf{W}_i^{T}\mathbf{f}}}$. Cross-entropy loss makes the predicted logit values to approximate to the ground truth. It encourages the differences between the largest logit and all others to become large, and this, combined with the bounded gradient reduces the ability of the model to adapt, resulting in a model too confident about its predictions. This, in turn, can lead to over-fitting. To build a robust model that can generalize well, \textbf{Label Smoothing} is proposed by Google Brain to address the problem. It encourages the activations of the penultimate layer to be close to the template of the correct class and equally distant to the templates of the incorrect classes. So the ground truth label $\mathbf{y}$ in cross-entropy loss can be defined as $y_i(j=c)=1-\delta$ and $y_i(j\neq c)=\frac{\delta}{C-1}$.

\noindent\textbf{Arcface loss} \cite{deng2019arcface} maps cartesian coordinates to spherical coordinates. It transforms the logit as $\mathbf{W}_i^{T}\mathbf{f}=\|\mathbf{W}_i\|\|\mathbf{f}\|\cos{\theta_i}$, where $\theta_i$ is the angle between the weight $\mathbf{W}_i$ and the feature $\mathbf{f}$. It fixes the individual weight $\|\mathbf{W}_i\|=1$ by $l_2$ normalisation and also fixes the embedding feature $\mathbf{f}$ by l2 normalisation and re-scale it to $s$, so $\hat{y_i}=\frac{e^{s\cos{\theta}_i}}{\sum_{i=1}^{C}e^{s\cos{\theta}_i}}$. To simultaneously enhace the intra-class compactness and inter-class discrepancy, Arcface adds an additive angular margin penalty $m$ in the intra-class measure. So $\hat{y_i}$ can rewritten as $\hat{y_i}=\frac{e^{s\cos(\theta_i+m)}}{e^{s\cos(\theta_i+m)}+\sum_{i=1,i\neq c}^{C-1}e^{s\cos{\theta}_i}}$.

\noindent\textbf{Circle loss}. The derivation process of circle loss is not described here in detail, it can refer to \cite{sun2020circle}.

\noindent\textbf{Triplet loss} ensures that an image of a specific person is closer to all other images of the same person than  to any images of other persons, which wants to make an image $x_i^{a}$ (anchor) of a specific person  closer to all other images $x_i^{p}$ (positive) of the same person than to any image $x_i^{n}$ (negative) of any other person in the image embedding space. Thus, we want $D(x_i^{a}, x_i^{p})+m<D(x_i^{a}, x_i^{n})$, where $D(:,:)$ is measure distance between a pair of person images. Then the \emph{Triplet Loss} with $N$ samples is defined as $\sum_{i=1}^{N}[m+D(g_i^{a},g_i^{p})-D(g_i^{a},g_i^{n})]$, where $m$ is a margin that is enforced between a pair of positive and negative.

\subsection{Training Strategy}
\begin{figure}[htp]
    \centering
       \vspace{0em}
    \includegraphics[width=8.5cm]{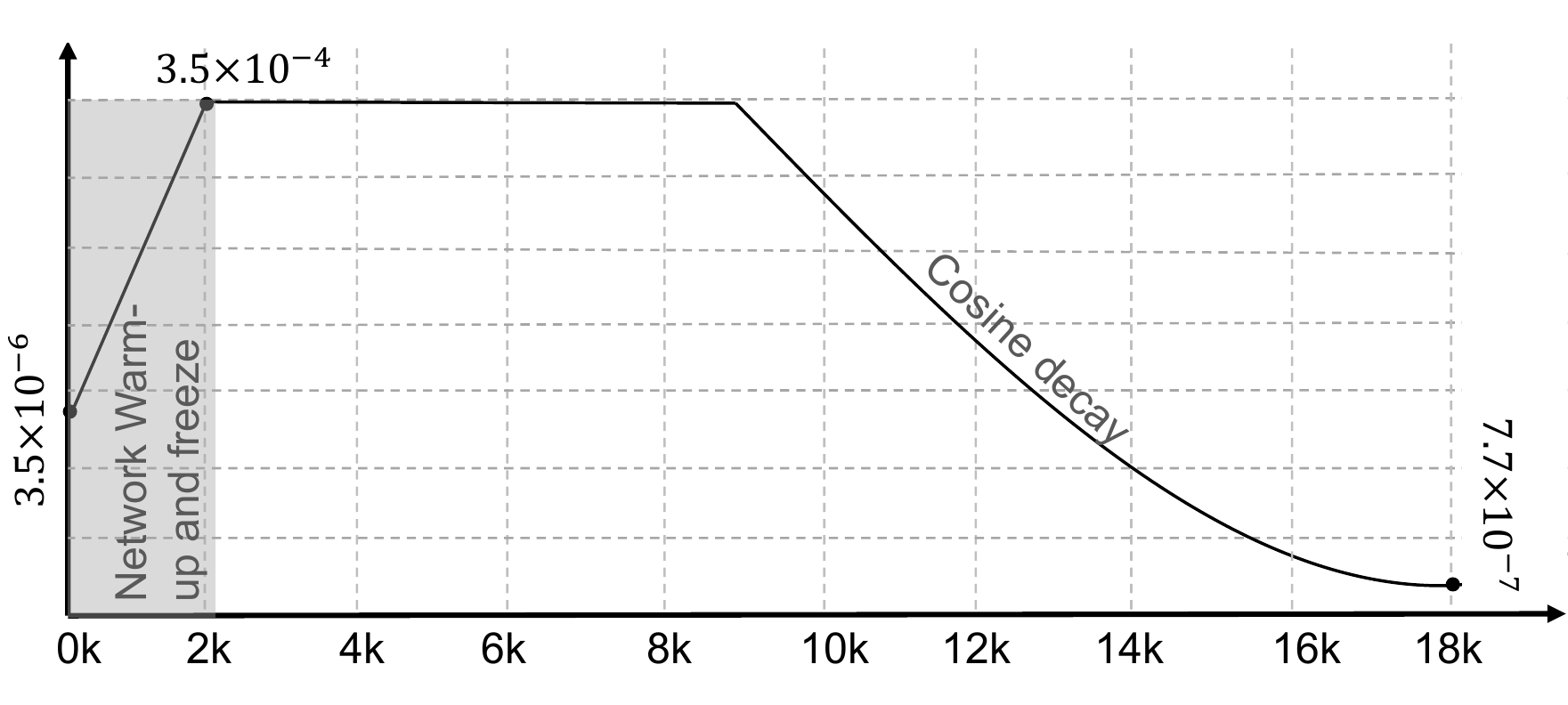}
     \caption{Learning rate curve as a function of the number of iteration}
    \label{fig5}
\end{figure}

Fig.~\ref{fig5} shows the train strategy that contains many tricks including learning rate for different iteration, network warm-up and freeze.

\noindent\textbf{Learning rate warm-up} helps to slow down the premature over-fitting of the mini-batch in the initial stage of the model training. Also, it helps to maintain the stability of the deep layer of the model. Therefore, we will give a very small learning rate, e.g., $3.5\times 10^{-5}$ in the initial training and then gradually increase it during the 2k iterations. After that, the learning rate remains at $3.5^{-4}$ between 2k iterations and 9k iterations. Then, the learning rate starts from $3.5\times 10^{-4}$ and decays to $7.7\times 10^{-7}$ at cosine rule after 9k iterations, the training is finished at 18k iterations.

\noindent\textbf{Backbone freeze.} To re-train a classification network to meet the requirement of our tasks, we use the collected data from the tasks to fine-tune on the ImageNet pre-trained model. Generally, we add a classifier that collected the network such as ResNet, and the classifier parameters are randomly initialized. To initialize the parameters of the classifier better, we only train the classifier parameters while freezing the network parameters without updating at the beginning of the training (2k iterations). After 2k iterations, we will free the network parameter for end-to-end training.

\section{Testing}
\subsection{Distance Metric.} Eucildean and cosine measure are implemented in FastReID. And we also implement a local matching method: deep spatial reconstruction (DSR).

\noindent\textbf{Deep spatial reconstruction.} Suppose there is a pair of person images $x$ and $y$. Denote the spatial features map from backbone as $\mathbf{x}$ for  $x$ with dimension dimension $w_x\times h_x \times d$,  and $\mathbf{y}$ for  $y$  with dimension $w_y\times h_y \times d$. The total $N$ spatial features from $N$ locations are aggregated into a matrix  $\mathbf{X}=[\mathbf{x}_n]_{n=1}^{N}\in  \mathbb{R}^{d\times N}$, where $N=w_x\times h_x$. Likewise, we construct the gallery feature matrix $\mathbf{Y}=\{\mathbf{y}_m\}_{m=1}^{M}\in \mathbb{R}^{d\times M}$, $M = w_y\times h_y$. Then, $\mathbf{x}_n$ can find the most similar spatial feature in $\mathbf{Y}$ to match, and its matching score $s_n$. Therefore, we try to obtain the similar scores for all spatial features of $\mathbf{X}$ with respect to $\mathbf{Y}$, and the final matching score can be defined as $s = \sum_{n=1}^{N} s_n$.

\subsection{Post-processing.} Two re-rank methods: K-reciprocal coding \cite{zhong2017re} and Query Expansion (QE) \cite{bhagwan2004total} are implemented in FastReID.

\noindent\textbf{Query expansion.} Given a query image, and use it to find $m$ similar gallery images. The query feature is defined as $\mathbf{f}_q$ and $m$ similar gallery features are defined as $\mathbf{f}_g$. Then the new query feature is constructed by averaging the verified gallery features and the query feature. So the new query feature $\mathbf{f}_{newq}$ can be defined as
\begin{equation}
 \mathbf{f}_{q_{new}} = \frac{\mathbf{f}_q+\sum_{i=1}^{m}\mathbf{f}_g^{(i)}}{m+1}.
\end{equation}
After that the new query feature $\mathbf{f}_{q_{new}}$ is used for following image retrieve. QE can be easily used for practical scenarios.

\subsection{Evaluation}
For performance evaluation, we employ the standard metrics as in most person re-identification literature, namely the cumulative matching cure (CMC) and the mean Average Precision (mAP). Besides, we also add two metrics: receiver operating characteristic (ROC) curve and mean inverse negative penalty (mINP) \cite{ye2020deep}.

\subsection{Visualization}
We provide a rank list tool of retrieval result that contributes to checking the problems of our algorithm that we haven't solved.

\section{Deployment}
In general, the deeper the model, the better the performance. However, too deep a model is not easy to deploy in edge computing hardware and AI chips since 1) it needs time-consuming inference; 2) many layers are difficult to implement on AI chips. Considering these reasons, we implement the knowledge distillation module in FastReID to achieve a high-precision, high-efficiency lightweight model. 

\begin{figure}[t]
    \centering
       \vspace{0em}
    \includegraphics[width=7cm]{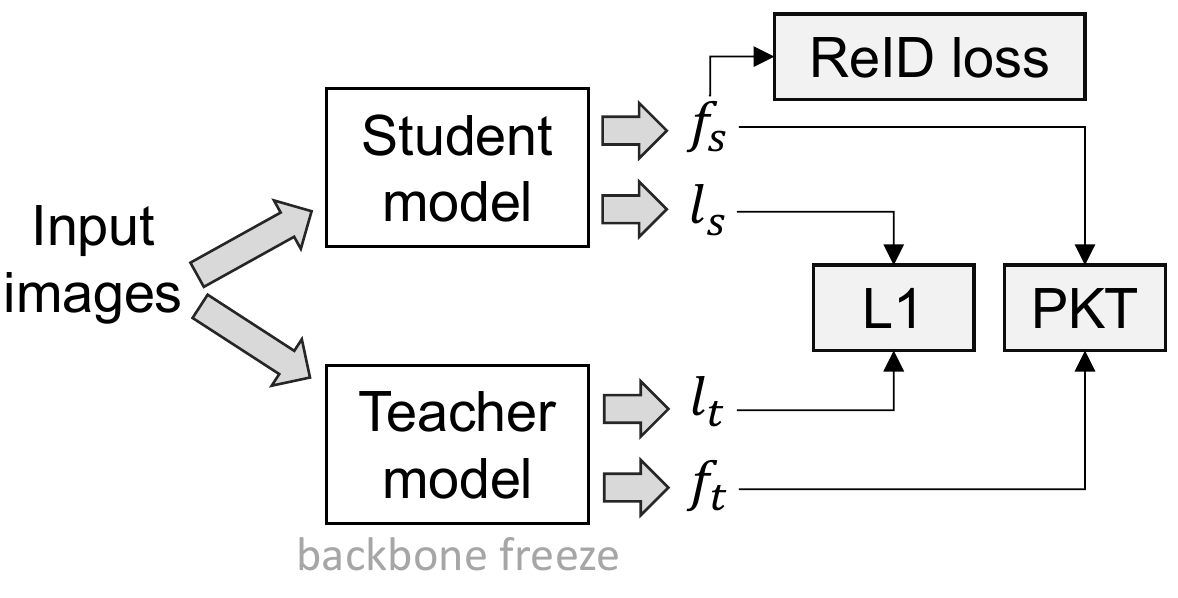}
     \caption{Illustration of knowledge distillation module}
    \label{fig6}
\end{figure}
As shown in Fig.~\ref{fig6}, given a pre-trained student model and a pre-trained teacher model on reid datasets, the teacher model is a deeper model with non-local module, ibn module and some useful tricks. The student model is simple and shallow. We adopt two-stream way to train the student model with teacher backbone freezing. The student and teacher models respectively output classifier logits $\mathbf{l}_s$, $\mathbf{l}_t$ and features $\mathbf{f}_s$, $\mathbf{f}_t$. We want student model to learn classification ability as much as possible about the teacher model, the logit learning can be defined as
\begin{equation}
\begin{array}{l}
 \displaystyle 
 \mathcal{L}_{logit} = \|\mathbf{l}_s-\mathbf{l}_t\|_1.
 \end{array}
\end{equation}

\begin{table}[t]
  \centering
  \fontsize{6.5}{7}\selectfont
  \caption{Performance comparison on Market1501, DukeMTMC and MSMT17 datasets.}
  \vspace{-2em}
  \label{tab1}
  \begin{center}
    \begin{tabular}{|l|cc|cc|cc|}
    \hline
    \multicolumn{1}{|c|}{\multirow{2}{*}{Methods}} &
    \multicolumn{2}{|c|}{\textbf{Market1501}}&\multicolumn{2}{|c|}{\textbf{DukeMTMC}}&\multicolumn{2}{|c|}{\textbf{MSMT17}} \cr \cline{2-7}
     \multicolumn{1}{|c|}{~}& R1 &mAP &R1 &mAP &R1 &mAP   \cr \hline
    SPReID \cite{Kalayeh_2018_CVPR} (CVPR'18)  & 92.5 & 81.3 & 84.4 & 70.1 &- &- \cr
    PCB \cite{sun2018beyond}  (ECCV'18)  &  92.3 & 77.4 & 81.8 & 66.1 &- &- \cr
    AANet \cite{Tay_2019_CVPR} (CVPR'19)   &93.9&83.4&87.7 &74.3 &-&-\cr
    IANet \cite{Hou_2019_CVPR} (CVPR'19)   &94.4 & 83.1&87.1 &73.4 &75.5 & 45.8\cr 
    CAMA \cite{Yang_2019_CVPR} (CVPR'19)   & 94.7 & 84.5 & 85.8 & 72.9  &- &- \cr
    DGNet \cite{Zheng_2019_CVPR} (CVPR'19)   & 94.8 & 86.0 & 86.6 &74.8 &- &- \cr
    DSAP \cite{Zhang_2019_CVPR} (CVPR'19)   & 95.7 & 87.6 & 86.2 & 74.3 &- &- \cr
    Pyramid \cite{Zheng_2019_CVPR} (CVPR'19)   &95.7&88.2&89.0 &79.0 &-&-\cr 
    Auto-ReID \cite{Quan_2019_ICCV} (ICCV'19) &94.5 & 85.1& - & - &78.2 & 52.5\cr 
    OSNet \cite{Zhou_2019_ICCV} (ICCV'19) &94.8 & 84.9&88.6 & 73.5 &78.7 &52.9\cr 
    MHN \cite{Chen_2019_ICCV} (ICCV'19) &95.1 & 85.0 & 89.1 & 77.2 & - & -\cr 
    $P^{2}$-Net \cite{Guo_2019_ICCV} (ICCV'19)  &95.2 & 85.6 &86.5 &75.1 &- & -\cr 
    BDB \cite{Dai_2019_ICCV} (ICCV'19)  &95.3 & 86.7& 89.0 & 76.0 &- & -\cr 
    FPR \cite{He_2019_ICCV} (ICCV'19) &95.4 & 86.6& 88.6 & 78.4 &- & -\cr
    ABDNet \cite{Chen_2019_ICCV} (ICCV'19)  &95.6 & 88.3& 89.0 & 78.6 &82.3 & 60.8\cr
    SONA \cite{Xia_2019_ICCV} (ICCV'19) &95.7 & 88.7& 89.3 & 78.1 &- & -\cr
    SCAL \cite{Chen_2019_ICCV} (ICCV'19) &95.8 & 89.3&  89.0 & 79.6 &- & -\cr 
    CAR \cite{Zhou_2019_ICCV} (ICCV'19) &96.1 & 84.7 & 86.3  & 73.1 &- & -\cr
    Circle Loss \cite{sun2020circle} (CVPR'20)  & 96.1 & 87.4 &  - & - &76.9 & 52.1\cr\hline
    FastReID (ResNet50) &  95.4 &  88.2 &  89.6 &  79.8 &  83.3 & 59.9\cr
    FastReID (ResNet50-ibn) &  95.7 &  89.3 &  91.3 &  81.6 &  84.0 & 61.2\cr
    FastReID (ResNeSt) &  95.0 &  87.0 &  90.5 &  79.1 &  82.6 & 58.2\cr
    
    FastReID-MGN (ResNet50-ibn) &  95.7 &  89.7 & 91.6 &  82.1 &  \bf 85.1 & \bf 65.4\cr\hline
    FastReID (ResNet101-ibn) & \bf 96.3 &  \bf 90.3 &\bf 92.4 & \bf 83.2 & \bf 85.1 & 63.3\cr
    + QE & \bf 96.5&  \bf 94.4 &\bf 93.4 & \bf 90.1 & \bf 87.9 & \bf 76.9\cr
    + Rerank & \bf 96.8 &  \bf 95.3 &\bf 94.4 & \bf 92.2 & \bf - & \bf -\cr
    \hline
    \end{tabular}
    \end{center}
\end{table}
\begin{table*}[t]
\caption{Ablation Studies of FastReID on DukeMTMC. (ResNet50, 384$\times$128).}
\label{tab:table4}
\centering
  \fontsize{8}{8}\selectfont
  \label{tab4}
\begin{tabular}{|cccccccccccc|} \hline
    \multirow{2}{*}{Bag-of-Tricks} & \multirow{2}{*}{IBN} & Auto- & Soft & Non- & Gem & Circle & Backbone & Cosine Lr& \multirow{2}{*}{R1}& \multirow{2}{*}{mAP}& \multirow{2}{*}{mINP}\\ 
      &  & Augment & Margin & Local & Pooling &Loss & Freeze &Scheduler & & & \\ \hline
$\surd$& & & & & & & & & 85.5& 75.2 &37.9\\ 
$\surd$&$\surd$ & & & & & & & & 89.2 &79.1 &43.9\\ 
$\surd$& & $\surd$& & & & & & & 84.9 &72.8& 34.5\\ 
$\surd$& & & $\surd$& & & & & & 86.1 &76.3 &39.0\\ 
$\surd$& & & & $\surd$& & & & & 87.3 &77.6 &42.0\\ 
$\surd$& & & & & $\surd$& & & & 87.4 &77.1 &40.3\\ 
$\surd$& & & & & & $\surd$& & & 88.7 &78.3 &41.8\\ 
$\surd$& & & & & & & $\surd$& &  85.9& 74.7& 36.4\\ 
$\surd$& & & & & & $\surd$& & $\surd$&  88.8 &77.8 &40.3\\ 
$\surd$& & & & & & $\surd$&$\surd$ & &  89.5& 78.3 &41.6\\ 
$\surd$& & & & & & $\surd$&$\surd$ &$\surd$ &  89.5 & 78.5& 42.5\\ 
$\surd$&$\surd$ &$\surd$ &$\surd$ &$\surd$ &$\surd$ & $\surd$&$\surd$&$\surd$&  91.3 &81.6& 47.6\\ \hline
\end{tabular}
\end{table*}

In order to ensure the consistency of student model and teacher model in the feature space distribution, probabilistic knowledge transfer model based on Kullback-Leibler divergence is used for optimizing the student model:
\begin{equation}
\left\{\begin{aligned}    
 \mathcal{L}_{PKT} & = \sum_{i=1}^{N}\sum_{j=1,i\neq j}^{N} p_{j|i} \log(\frac{p_{j|i}}{p_{i|j}})  \\
 p_{i|j} &= \frac{K(\mathbf{f}_s^{i}, \mathbf{f}_s^{j})}{\sum_{j=1,i\neq j}^{N}K(\mathbf{f}_s^{i}, \mathbf{f}_s^{j})}  \\
 p_{j|i} &= \frac{K(\mathbf{f}_t^{i}, \mathbf{f}_t^{j})}{\sum_{j=1,i\neq j}^{N}K(\mathbf{f}_t^{i), \mathbf{f}_t^{j}}}
 \end{aligned}\right.
\end{equation}
where $K(:,:)$ is cosine similarity measure.

At the same time, the student model needs ReID loss $\mathcal{L}_{reid}$ to optimize the entire network. Therefore, the total loss is:
\begin{equation}
\begin{array}{l}
 \displaystyle 
 \mathcal{L}_{kd} = \mathcal{L}_{logit} + \alpha\mathcal{L}_{PKT}+\mathcal{L}_{reid}.
 \end{array}
\end{equation}
After finish training, the $\mathbf{f}_s$ is used for inference. 

We also provide model conversion tool (\textbf{PyTorch} $\to$ \textbf{Caffe} and \textbf{PyTorch} $\to$ \textbf{TensorRT})  in the FastReID library.

\section{Projects}
\subsection{Person Re-identification}
\noindent\textbf{Datasets.} Three person re-id benchmarking datasets: Market1501 \cite{bai2017scalable}, DukeMTMC \cite{zheng2017unlabeled}, MSMT17 \cite{qian2019leader} are used for evaluating the FastReID. We won't go into the details of the database here.

\begin{figure*}[t]
    \includegraphics[width=17cm]{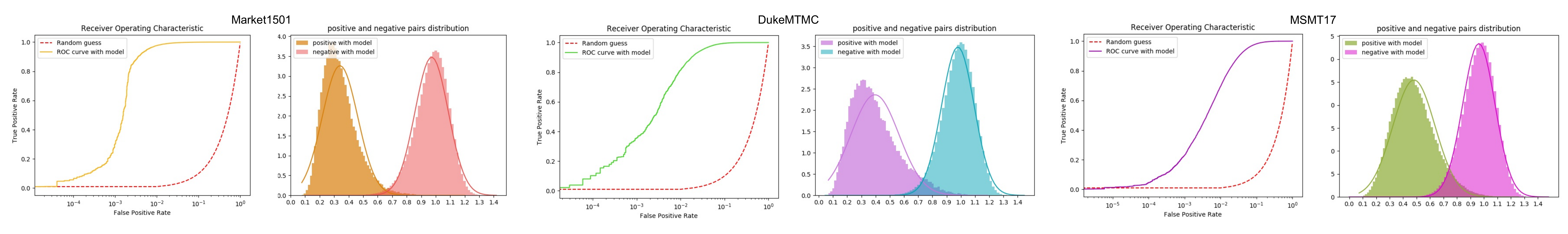}
     \caption{ROC curves and distribution curves between intra-class and inter-class samples on three benchmarking datasets for FastReID (ResNet101-ibn)}
    \label{fig7}
    \vspace{-1em}
\end{figure*}

\noindent\textbf{FastReID Setting.} We use flipping, random erasing and auto-augment to process the training image. The IBN-ResNet101 with a Non-local module is used as the backbone. The gem pooling and bnneck are used as the head and aggregation layer, respectively. For the batch hard triplet loss
function, one batch consists of 4 subjects, and each subject has 16 different images, and we use circle loss and triplet loss to train the whole network. 

\noindent\textbf{Result.} The state-of-the-art algorithms published in CVPR, ICCV, ECCV during 2018-2020 are listed in Table~\ref{tab1}, FastReID achieves the best performance on Market1501  96.3\%(90.3\%), DukeMTMC 92.4\%(83.2\%) and MSMT17 85.1\%(65.4\%) at rank-1/mAP accuracy, respectively. Fig.~\ref{fig7} shows the ROC curves on the three benchmarking datasets.

\subsection{Cross-domain Person Re-identification}
\noindent\textbf{Problem definition.} Cross-domain person re-identification aims at adapting the model trained on a labeled source domain dataset to another target domain dataset without any annotation.

\begin{figure}[t]
    \centering
    \includegraphics[width=8.3cm]{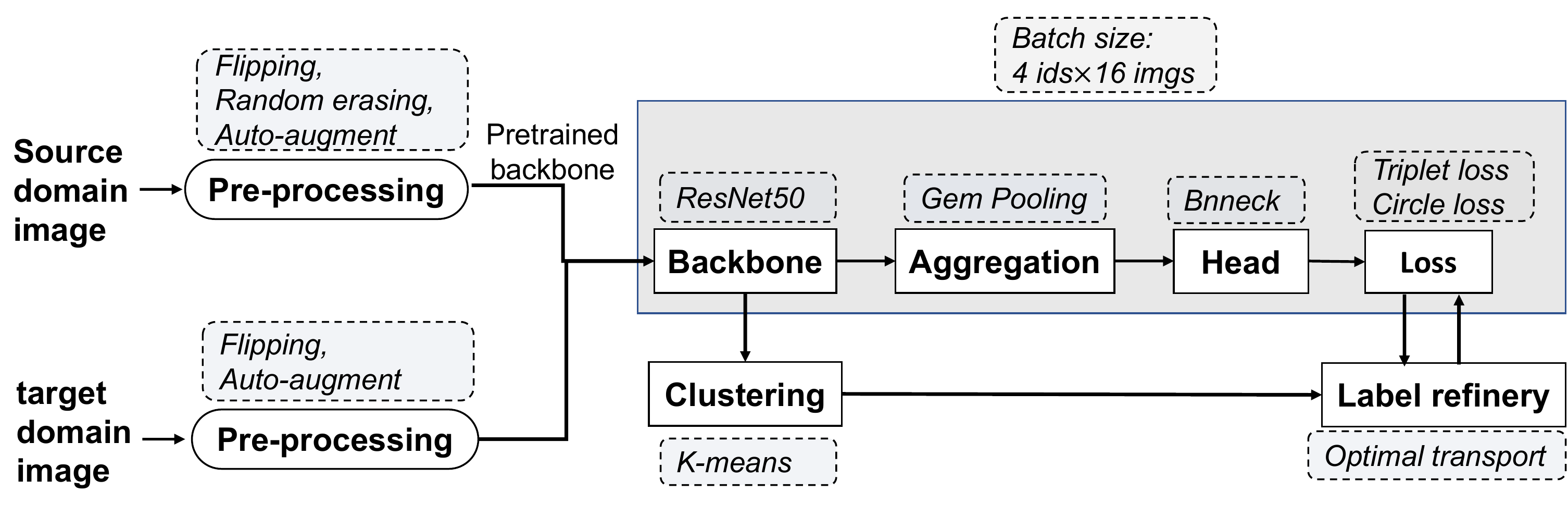}
     \caption{{Framework of FastReID-MLT}}
    \label{fig9}
\end{figure}

\noindent\textbf{Setting.} We propose a cross-domain method FastReID-MLT that adopts mixture label transport to learn pseudo label by multi-granularity strategy. We first train a model with a source-domain dataset and then finetune on the pre-trained model with pseudo labels of the target-domain dataset. FastReID-MLT is implemented by ResNet50 backbone, gem pooling and bnneck head. For the batch hard triplet loss function, one batch consists of 4 subjects, and each subject has 16 different images, and we use circle loss and triplet loss to train the whole network. 
Detailed configuration can be found on the GitHub website. The framework of FastReID-MLT is shown in Fig.~\ref{fig9}. 
\begin{table}[t]
        
\fontsize{8}{8}\selectfont
\centering
\caption{Performance comparison to the unsupervised cross-domain re-id SOTA methods on three benchmark datasets. ``BOT'' denotes to the bag of tricks method, which is a strong baseline in the ReID task. M: Market1501, D: DukeMTMC, MS: MSMT17.}
\vspace{-0.5em}
        \begin{center}
            \begin{tabular}{|l|cc|cc|}
                \hline
                \multicolumn{1}{|c|}{\multirow{2}{*}{Methods}} & \multicolumn{2}{c|}{\scriptsize{D$\to$M}} & \multicolumn{2}{c|}{M$\to$D} \\
                \cline{2-5}
                \multicolumn{1}{|c|}{} & mAP & R1  & mAP & R1  \\ 
                \hline 
                TJ-AIDL~\cite{wang2018transferable} (CVPR'18) & 26.5 & 58.2  & 23.0 & 44.3  \\
                SPGAN~\cite{deng2018image} (CVPR'18) & 22.8 & 51.5  & 22.3 & 41.1  \\
                ATNet~\cite{Liu2019cvpr}(CVPR'19) & 25.6& 55.7&  24.9 &45.1 \\
                SPGAN+LMP~\cite{DengWeijian2018cvpr}(CVPR'18) &26.7 &57.7 & 26.2&46.4  \\
                HHL~\cite{zhong2018generalizing} (ECCV'18) & 31.4 & 62.2  & 27.2 & 46.9  \\

                ARN~\cite{li2018adaptation} (CVPR'18-WS) & 39.4 & 70.3 & 33.4 & 60.2  \\
                ECN~\cite{zhong2019invariance} (CVPR'19) & 43.0 & 75.1  & 40.4 & 63.3  \\
                UCDA~\cite{qi2019novel} (ICCV'19) & 30.9 & 60.4  & 31.0 & 47.7  \\
                PDA-Net~\cite{li2019cross} (ICCV'19) & 47.6 & 75.2  & 45.1 & 63.2  \\
                PCB-PAST~\cite{zhang2019self} (ICCV'19) & 54.6 & 78.4  & 54.3 & 72.4  \\
                SSG~\cite{yang2019selfsimilarity} (ICCV'19) & 58.3 & 80.0  & 53.4 & 73.0  \\
                MPLP+MMCL~\cite{WANG2020cvpr1} (CVPR'20) & 60.4 &84.4  & 51.4&72.4   \\ 
                AD-Cluster~\cite{zhai2020adcluster} (CVPR'20)& {68.3} & {86.7}  & {54.1} & {72.6}  \\ 
                MMT~\cite{ge2020mutual} (ICLR'20) & {71.2} & {87.7}  & {65.1} & {78.0}  \\
                \hline
                FastReID-MLT  & \textbf{80.5} & \textbf{92.7} & \textbf{69.2}  & \textbf{82.7}  \\
                
                Supervised learning (BOT~\cite{Luo2019CVPRWorkshops})&85.7&94.1 &75.8&86.2   \\\hline
                \hline

                \multicolumn{1}{|c|}{\multirow{2}{*}{Methods}} & \multicolumn{2}{c|}{M$\to$MS} & \multicolumn{2}{c|}{D$\to$MS} \\
                \cline{2-5}
                \multicolumn{1}{|c|}{} & mAP & R1 & mAP & R1  \\ 
                \hline 
                PTGAN~\cite{wei2018person} (CVPR'18) & 2.9 & 10.2  & 3.3 & 11.8  \\    
                ENC~\cite{zhong2019invariance} (CVPR'19) & 8.5 & 25.3  & 10.2 & 30.2  \\
                SSG~\cite{yang2019selfsimilarity} (ICCV'19) & 13.2 & 31.6 & 13.3 & 32.2  \\
                DAAM~\cite{Huang2020aaai} (AAAI'20)& 20.8 &44.5  & 21.6 & 46.7  \\
                
                MMT~\cite{ge2020mutual} (ICLR'20) & 22.9 & 49.2 &  23.3 & 50.1  \\
                \hline
                FastReID-MLT & \textbf{26.5} & \textbf{56.6}  & \textbf{27.7} & \textbf{59.5}  \\
                Supervised learning (BOT~\cite{Luo2019CVPRWorkshops})&48.3&72.3 &48.3&72.3   \\
                \hline
            \end{tabular}
        \end{center}
        \label{tab:sota}
            \vspace{-10pt}
    \end{table}
    
\noindent\textbf{Result.} Table~\ref{tab:sota} shows the results on several datasets, FastReID-MLT can achieve 92.7\%(80.5\%), 82.7\%(69.2\%) under D$\to$M, M$\to$D settings. The result is close to supervised learning results.

\subsection{Partial Person Re-identification}
\noindent\textbf{Problem definition.} Partial person re-identification (re-id) is a challenging problem, where only several partial observations (images) of people are available for matching.

\begin{figure}[htp]
    \centering
    \includegraphics[width=8.4cm]{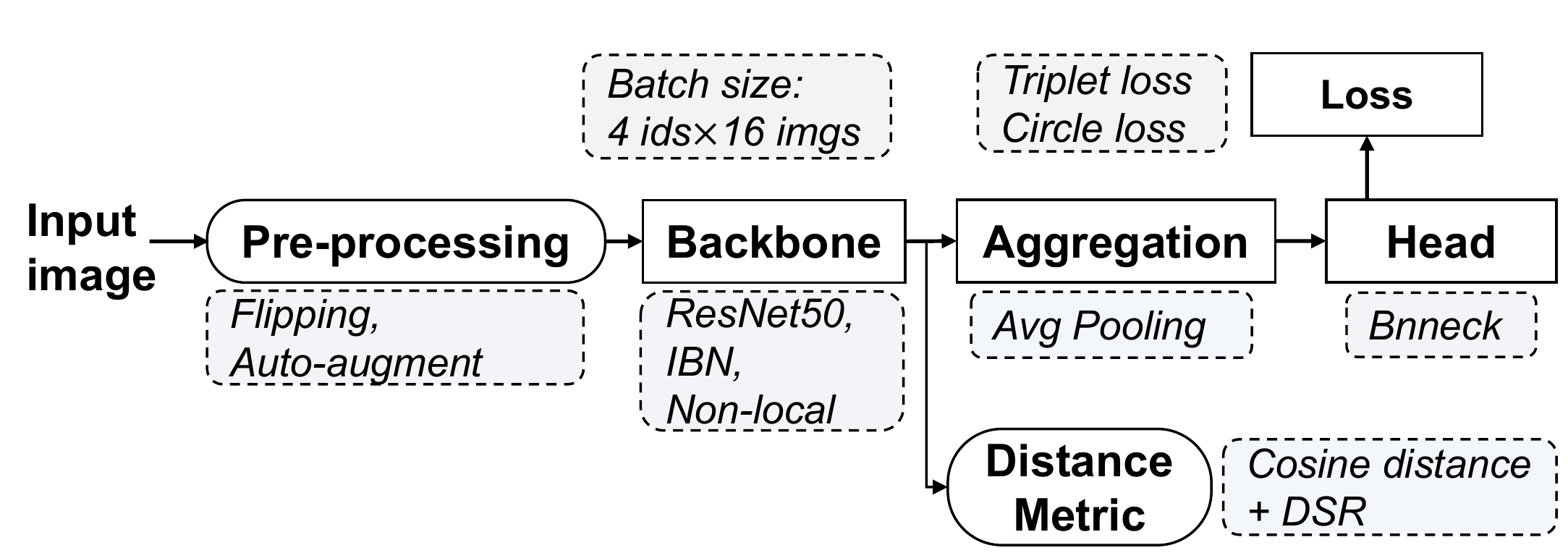}
     \caption{{Framework of FastReID-DSR}}
    \label{fig10}
\end{figure}

\noindent\textbf{Setting.} The setting as shown in Fig.~\ref{fig10}.

\begin{table}[t]
\caption{Comparison of the state-of-the-art Partial Person Re-ID methods on the PartialREID, OccludedREID and PartialiLIDS datastes.}
\label{tab5}
\centering
  \fontsize{6.7}{7.5}\selectfont
  \label{tab5}
\begin{tabular}{|l|cc|cc|cc|} \hline
    \multicolumn{1}{|c|}{\multirow{2}{*}{Methods}} & \multicolumn{2}{c|}{PartialREID} & \multicolumn{2}{c|}{OccludedREID} & \multicolumn{2}{c|}{PartialiLIDS} \\ \cline{2-7}
    \multicolumn{1}{|c|}{} & R1 & mAP & R1 & mAP& R1 & mAP \\ \hline 

PCB~\cite{sun2018beyond} (ECCV'18)              & 56.3& 54.7 &41.3  & 38.9& 46.8& 40.2     \\
SCPNet~\cite{fan2018scpnet} (ACCV'18)  & 68.3& - &-  & -  & -& -     \\
DSR~\cite{he2018deep} (CVPR'18)         & 73.7 & 68.1 & 72.8  & 62.8 & 64.3& 58.1     \\
VPM~\cite{sun2019perceive} (CVPR'19)       & 67.7 & -&    - & -     & 65.5 & -     \\ 
FPR~\cite{he2019foreground}     (ICCV'19)             & 81.0 & 76.6&     78.3  & 68.0& 68.1& 61.8         \\
HOReID~\cite{wang2020high}    (CVPR'20)    & \bf 85.3 &-&     80.3 & 70.2    & 72.6 &-\\ \hline
FastReID-DSR                 & 82.7 &\textbf{76.8}& \textbf{81.6} & \textbf{70.9} &\textbf{73.1}&\textbf{79.8}     \\  \hline
\end{tabular}
\end{table}
\begin{figure}[t]
    \centering
    \includegraphics[width=8.5cm]{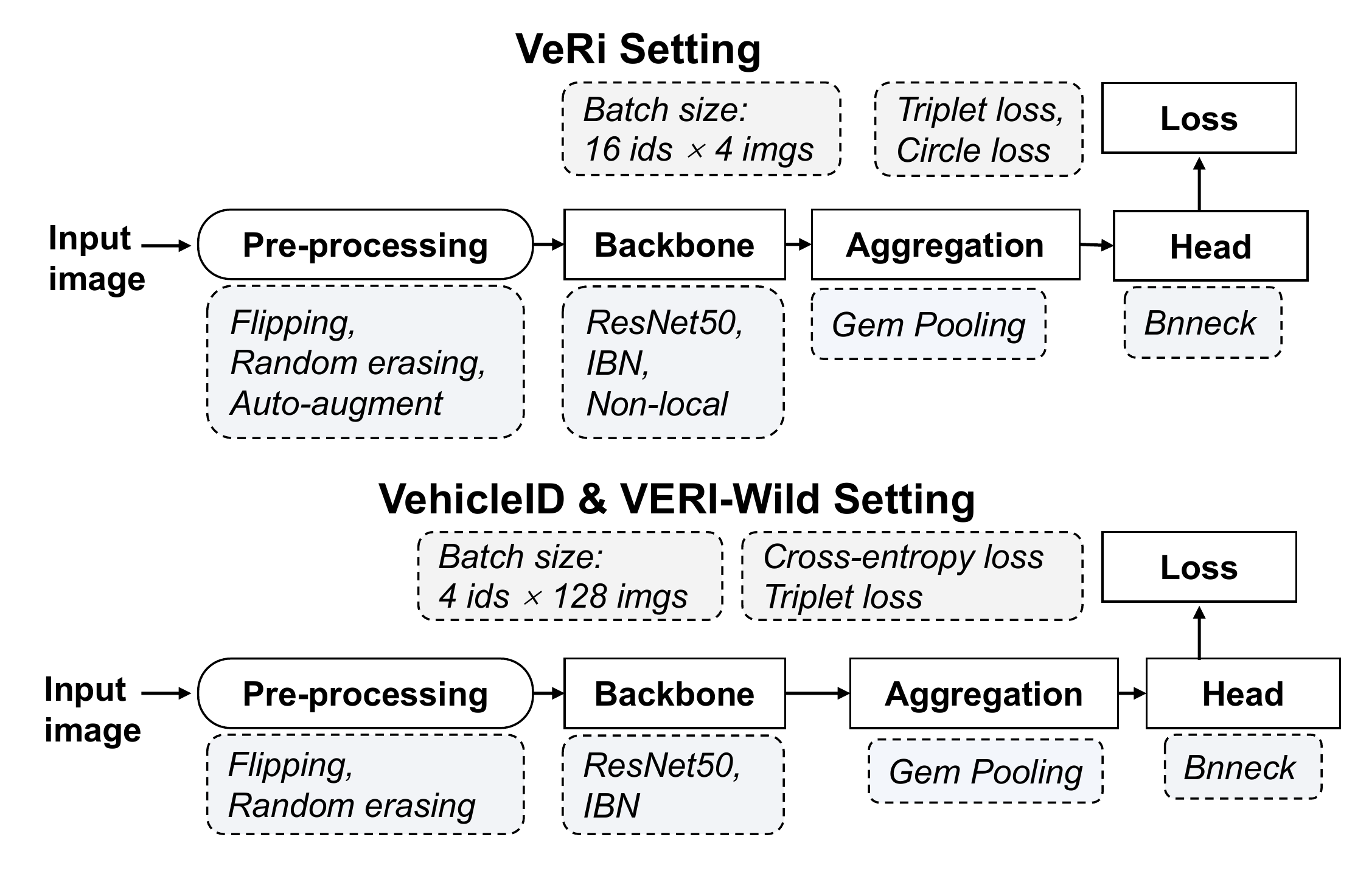}
    \caption{{Framework of FastReID on VehicleID and VERI-Wild}}
    \label{fig11}
\end{figure}

\noindent\textbf{Result.} Table~\ref{tab5} shows the results on PartialREID, OccludedREID and PartialiLIDS datasets. FastReID-DSR can achieve 82.7\% (76.8\%), 81.6\% (70.9\%) and 73.1\% (79.8) at rank-1/mAP metrics.

\subsection{Vehicle Re-identification}
\noindent\textbf{Datasets.} Three vehicle re-id benchmarking datasets: VeRi, VehicleID and VERI-Wild are used for evaluating the proposed FastReIDin the FastReID. We won't go into the details of the database here.

\noindent\textbf{Settings.} The setting as shown in Fig.~\ref{fig11}.

\noindent\textbf{Result.} The state-of-the-art algorithms published during 2015-2019 are listed in Table~\ref{tab5}, Table~\ref{tab6}, Table~\ref{tab7}. FastReID achieves the best performance on VeRi, VehicleID and VERI-Wild, respectively.

\section{Conclusion}
This paper introduces a open source library namely FastReID for general instance re-identification. Experimental results demonstrated the versatility and effectiveness of FastReID on multiple tasks, such as person re-identification and vehicle re-identification. We’re sharing FastReID because open source research platforms are critical to the rapid advances in AI made by the entire community, including researchers and practitioners in academia and industry. We hope that releasing FastReID will continue to accelerate progress in the area of general instance re-identification. We also look forward to collaborating with learning from each other for advancing the development of computer vision.

\begin{table}[t]
\caption{Comparison of the state-of-the-art vehicle Re-Id methods on the VeRi dataset.}
\label{tab:table4}
\centering
  \fontsize{8}{8}\selectfont
  \label{tab5}
\begin{tabular}{|l|c|c|c|} \hline
Methods                                           & mAP (\%)     & R-1 (\%)     & R-5 (\%)  \\  \hline

Siamese-CNN~\cite{iccv/ShenXLYW17}    (ICCV'17)    & 54.2         & 79.3             & 88.9     \\
FDA-Net~\cite{cvpr/LouB0WD19}    (CVPR'19)    & 55.5         & 84.3             & 92.4     \\ 
Siamese-CNN+ST~\cite{iccv/ShenXLYW17} (ICCV'17)    & 58.3           & 83.5             & 90.0     \\ 
PROVID~\cite{tmm/LiuLMM18}    (TMM'18)    & 53.4         & 81.6             & 95.1     \\ 
PRN~\cite{cvpr/HeLZT19}        (CVPR'19)    & 70.2         & 92.2             & 97.9     \\
PAMTRI~\cite{iccv/TangNBTHKWY19}(ICCV'19)    & 71.8         & 92.9             & 97.0     \\  
PRN~\cite{cvpr/HeLZT19}    (CVPR'19)    & 74.3         & 94.3             & 98.9     \\ \hline
FastReID                                 & \textbf{81.9} & \textbf{97.0} & \textbf{99.0}     \\  \hline
\end{tabular}
\end{table}

\begin{table}[t]
\caption{Comparison of the state-of-the-art vehicle Re-Id methods on the VehicleID dataset.}
\label{tab6}
\centering
  \fontsize{8}{8}\selectfont
\begin{tabular}{|l|cc|cc|cc|} \hline
    \multicolumn{1}{|c|}{\multirow{2}{*}{Methods}}        & \multicolumn{2}{c|}{Small}    & \multicolumn{2}{c|}{Medium}    & \multicolumn{2}{c|}{Large} \cr \cline{2-7}                             & R-1             & R-5          & R-1              & R-5         & R-1              & R-5    \\  \hline
DRDL~\cite{cvpr/LiuTWPH16}       & 48.9        & 73.5             & 42.8         & 66.8        & 38.2             & 61.6    \\
NuFACT~\cite{tmm/LiuLMM18}        & 48.9        & 69.5             & 43.6         & 65.3        & 38.6             & 60.7    \\
VAMI~\cite{cvpr/Zhou018}              & 63.1        & 83.3             & 52.9         & 75.1        & 47.3             & 70.3    \\
FDA-Net~\cite{cvpr/LouB0WD19}    & -            & -                 & 59.8         & 77.1        & 55.5             & 74.7    \\
AAVER~\cite{iccv/Khorramshahi0PR19}& 74.7        & 93.8             & 68.6         & 90.0        & 63.5             & 85.6    \\
OIFE~\cite{iccv/WangTLYYSYWLW17}& -        & -                 & -             & -            & 67.0             & 82.9    \\
PRN~\cite{cvpr/HeLZT19}            & 78.4        & 92.3         & 75.0             & 88.3        & 74.2             & 86.4     \\\hline
FastReID                         & \textbf{86.6} & \textbf{97.9} & \textbf{82.9} & \textbf{96.0} & \textbf{80.6} & \textbf{93.9} \\  \hline
\end{tabular}
\end{table}

\begin{table}[t]
\caption{Comparison of the state-of-the-art vehicle Re-Id methods on the VERI-Wild dataset.}
\label{tab7}
\centering
  \fontsize{8}{8}\selectfont
\begin{tabular}{|l|cc|cc|cc|} \hline
\multicolumn{1}{|c|}{\multirow{2}{*}{Methods}}                        & \multicolumn{2}{c|}{Small}    & \multicolumn{2}{c|}{Medium}    & \multicolumn{2}{c|}{Large} \cr \cline{2-7} 
                                  & mAP         & R-1              & mAP         & R-1         & mAP         & R-1    \\  \hline
GoogLeNet~\cite{cvpr/YangLLT15}  & 24.3        & 57.2             & 24.2         & 53.2        & 21.5             & 44.6    \\
DRDL~\cite{cvpr/LiuTWPH16}        & 22.5        & 57.0             & 19.3         & 51.9        & 14.8             & 44.6    \\
FDA-Net~\cite{cvpr/LouB0WD19}    & 35.1        & 64.0        & 29.8         & 57.8        & 22.8            & 49.4    \\
MLSL~\cite{access/AlfaslyHLLJLZ19}& 46.3        & 86.0        & 42.4         & 83.0        & 36.6            & 77.5    \\ \hline
FastReID                         & \textbf{87.7} & \textbf{96.4} & \textbf{83.5} & \textbf{95.1} & \textbf{77.3} & \textbf{92.5} \\  \hline
\end{tabular}
\end{table}

{
\balance
\small
\bibliographystyle{ieee_fullname}
\bibliography{egbib.bib}
}

\end{document}